\documentclass{llncs}

\usepackage{booktabs}
\usepackage{array}

\usepackage{amsmath, amssymb, mathtools}
\usepackage{bm}

\DeclareMathOperator*{\argmax}{argmax}

\usepackage{graphicx}
\graphicspath{{./images/}}

\usepackage{overpic}

\usepackage{color}

\usepackage{tikz}
\usetikzlibrary{arrows.meta, positioning}
\usetikzlibrary{arrows, shapes, backgrounds, fit} 
\usetikzlibrary{shadows, positioning}
\usepackage{circuitikz}
\usepackage{tikzscale}

\usepackage{pgfplots}
\newlength\figureheight 
\newlength\figurewidth

\usepackage[title]{appendix}

\usepackage{ctable}

\usepackage{cite}

\usepackage{contour}
\usepackage{multirow}

\usepackage{placeins}

\pgfplotsset{compat=1.14}

\begin{document}\sloppy

\title{3D Shape Segmentation\\ with Geometric Deep Learning}

\titlerunning{3D Shape Segmentation\\ with Geometric Deep Learning}

\author{Davide Boscaini \and Fabio Poiesi}
\authorrunning{D.~Boscaini and F.~Poiesi}

\institute{
Technologies of Vision, Fondazione Bruno Kessler\thanks{This research has been partially funded by the European Union's Horizon 2020 research and innovation programme under grant agreement number 687757.}\\
Via Sommarive 18, 38123, Trento, Italy\\
\email{<dboscaini,poiesi>@fbk.eu}
}

\maketitle 

\begin{abstract}
The semantic segmentation of 3D shapes with a high-density of vertices could be impractical due to large memory requirements. To make this problem computationally tractable, we propose a neural-network based approach that produces 3D augmented views of the 3D shape to solve the whole segmentation as sub-segmentation problems. 3D augmented views are obtained by projecting vertices and normals of a 3D shape onto 2D regular grids taken from different viewpoints around the shape. 
These 3D views are then processed by a Convolutional Neural Network to produce a probability distribution function (pdf) over the set of the semantic classes for each vertex. These pdfs are then re-projected on the original 3D shape and postprocessed using contextual information through Conditional Random Fields. We validate our approach using 3D shapes of publicly available datasets and of real objects that are reconstructed using photogrammetry techniques. 
We compare our approach against state-of-the-art alternatives.

\end{abstract}

\section{Introduction}

Traditional Convolutional Neural Networks (CNNs) use a cascade of learned convolution filters, pooling operations and activation functions to transform image data into feature embeddings processable by fully connected layers that classify the image content \cite{He2016}.
Typically, 3D deep-learning approaches extend traditional 2D methods to non-Euclidean domains as the convolution operation is not well defined in 3D \cite{Monti2017}.
One of the most challenging researched topic related to 3D deep learning is the semantic segmentation of 3D shapes as it is key to support computer graphics applications such as shape editing \cite{Yu2004} and modelling \cite{Chen2015}. Challenges to segment 3D shapes include dealing with different topologies, handling noisy geometries and different resolutions, and modeling semantic representations for different segments.

3D segmentation can be performed through multi-view \cite{Kalogerakis2017,Su2015}, volumetric \cite{Wu2015} or intrinsic \cite{Qi2017,Monti2017} deep learning-based approaches.
Multi-view and volumetric approaches use Euclidean structures, such as 2D or 3D grids, respectively, to process 3D shapes with 2D CNNs \cite{Kalogerakis2017, Su2015, Wu2015}. In particular, multi-view approaches simplify the representation of a 3D model using a set of rendered depth images taken from different viewpoints around the model, thus making the segmentation independent of the 3D-model polygon density \cite{Kalogerakis2017,Su2015}. Multi-view approaches cannot fully exploit the geometric properties of the 3D shape (e.g.~face normals) because geometric information can be lost when data are projected in 2D. Volumetric approaches approximate the 3D shape using voxels which could overshadow geometric details of the object \cite{Wu2015}.
Intrinsic approaches can be further divided into point-based and convolution-based approaches. Point-based approaches define feature extractors directly on the shape vertices \cite{Qi2017}, whereas convolution-based approaches extend the traditional convolution operations from grid-like structures to triangular meshes \cite{Monti2017}. Point-based approaches mostly process each vertex of the shape independently and loosely exploit local information \cite{Qi2017}. The additional structures used by conventional convolution-based approaches increase the shape representation complexity hence prohibiting the processing of high-density polygon models \cite{Monti2017}. Typically, 3D segmentation approaches validate their performance on datasets collected in controlled scenarios, and they mostly lack of an evaluation carried out on 3D models reconstructed using photogrammetric techniques \cite{Nocerino2017}.

In this paper we propose a novel 3D segmentation approach that retains both the advantages of view-based \cite{Kalogerakis2017} and intrinsic approaches \cite{Monti2017} by building 3D augmented views from multiple viewpoints around a 3D shape.
3D augmented views are a projection of 3D shape portions on 2D regular grids, where each cell of the grid encodes the information about depth and normal of the corresponding projected portion. 
This allows us to significantly reduce the number of parameters to learn and to perform 3D segmentation of shapes with diverse mesh topology (e.g.~polygon structure and/or density). 
We evaluate our approach on synthetic 3D shapes from publicly available datasets, and on 3D shapes of objects we captured with a smartphone and reconstructed using photogrammetry techniques.
Results show that the proposed approach can achieve state-of-the-art accuracy by using only $1\%$ of the parameters used by the alternative approaches.

\section{Our approach}

\subsection{Problem formulation}\label{sec:prob_formulation}

Given a 3D shape $\mathcal{X} \subset \mathbb{R}^3$ composed of vertices $x \in \mathcal{X}$, we design a neural-network based approach $\boldsymbol{p}(x) = \boldsymbol{\Gamma}_{\boldsymbol{\Theta}}(x)$ that outputs a probability distribution $\boldsymbol{p}(x)$ over the label space $\mathcal{L} = \{1, \dots, L\}$, where $L$ is the number of segmentation labels.
The output segmentation of $\mathcal{X}$ is computed as 
\begin{equation*}
    h(x) = \argmax_{\ell = 1, \dots, L} \boldsymbol{p}(x),
\end{equation*}
where $h(x)$ is a label defining the segment class of the vertex $x$.

\begin{figure}[t]
	\centering
	\begin{overpic}[width=\textwidth]{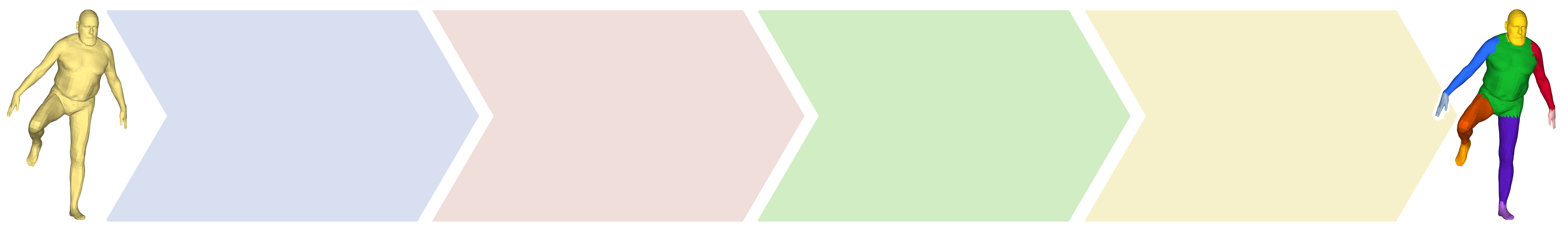} 
	    \put(1.5, 0){\contour{white}{\footnotesize $\mathcal{X}$}}
        \put(12, 10){\contour{white}{\footnotesize Shape}}
        \put(12, 7){\contour{white}{\footnotesize decomposition}}
        \put(12, 4){\contour{white}{\footnotesize $(\mathcal{V}^{(m)}, t^{(m)})$}}
        \put(33, 10){\contour{white}{\footnotesize Feature}}
        \put(33, 7){\contour{white}{\footnotesize extraction}}
        \put(33, 4){\contour{white}{\footnotesize \& class. $\boldsymbol{g}_{\mathcal{V}^{(m)}}$}}
        \put(54, 10){\contour{white}{\footnotesize Prediction}}
        \put(54, 7){\contour{white}{\footnotesize aggregation}}
        \put(54, 4){\contour{white}{\footnotesize $\boldsymbol{g}_{\mathcal{X}}$}}
        \put(75, 10){\contour{white}{\footnotesize Prediction}}
        \put(75, 7){\contour{white}{\footnotesize refinement}}
        \put(75, 4){\contour{white}{\footnotesize $\boldsymbol{p}_{\mathcal{X}}$}}
        \put(90.2, 0){\contour{white}{\footnotesize $\mathcal{X} \times \mathcal{L}$}}
    \end{overpic}
	\caption{Our approach outline. 3D augmented views from different viewpoints are computed from the 3D shape (shape decomposition). Point-wise features (i.e.~coordinates and surface normals) are extracted from these 3D views and classified to obtain segmentation predictions. Predictions are re-projected and aggregated on the original shape, and refined through a Conditional Random Field for local prediction consistency.}
	\label{fig:arch}
\end{figure}
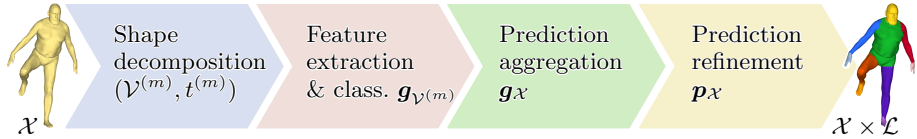

The neural network $\boldsymbol{\Gamma_{\boldsymbol{\Theta}}}$ can be defined as a a parametric function in the set of learnable parameters (i.e.~weights) $\boldsymbol{\Theta}$.
$\boldsymbol{\Gamma_{\boldsymbol{\Theta}}}$ is composed of four modules, namely shape decomposition, feature extraction and classification, feature aggregation and prediction refinement.
\emph{Shape decomposition} transforms the input 3D shapes into 3D augmented views, or \emph{3D views}.
Each 3D view is processed by a \emph{feature extraction and classification} network, namely \emph{ViewNet}, that predicts the class of each vertex. 
\emph{Prediction aggregation} re-projects the predictions of ViewNet of each 3D view onto the original 3D shape.
\emph{Prediction refinement} improves class prediction using contextual information on the original shape. Fig.~\ref{fig:arch} depicts the block diagram.

\subsection{Shape decomposition}

We simplify the 3D shape representation (e.g.~triangular meshes, quad meshes, CAD models) by decomposing the input shape into several components.
Shape decomposition can be performed by clustering shape vertices \cite{Hua2015}, by using geometrical primitives \cite{Kaiser2019}, or by generating range scans from different viewpoints \cite{Kalogerakis2017}.
We use a similar approach to the latter in order to process the 3D shape regardless its 3D representation, resolution and vertex topology.

Given $\mathcal{X}$ in the form of a triangular mesh with vertices $\boldsymbol{X} = (\boldsymbol{x}_1, \dots, \boldsymbol{x}_N)$, $\boldsymbol{x}_n \in \mathbb{R}^3$, $n=1, \dots, N$, we simplify $\mathcal{X}$ by building 3D views from $M$ different viewpoints.
Let $\mathcal{I}(u, v; \boldsymbol{w}_m) = (u, v, d(u, v))$ be a range scan that is captured from the $m$th viewpoint $\boldsymbol{w}_m$, where $(u, v)$ is the coordinate of a pixel, $d(u, v)$ is the depth value of the 3D shape, and $m = 1, \dots, M$.
Let $\mathcal{V}^{(m)}$ be the $m$th 3D view whose vertices $\boldsymbol{V}^{(m)} = (\boldsymbol{v}_1, \dots, \boldsymbol{v}_{N^{(m)}})$ are obtained by registering the coordinates $(u, v, d(u, v))$ of the range scan to the coordinates of the vertices $\boldsymbol{X}$. The faces of the 3D view are obtained by connecting depth values using the typical regular grid pattern of 2D images. For each vertex $\boldsymbol{v}_n \in \boldsymbol{V}^{(m)}$ we compute the surface normal $\boldsymbol{n}(\boldsymbol{v}_n) \in \mathbb{R}^3$ to define the signal on the 3D view as $\boldsymbol{f}(\boldsymbol{v}_n) = \left( \boldsymbol{v}_n, \boldsymbol{n}(\boldsymbol{v}_n) \right)$. 
The relation between a 3D view and the input 3D shape is defined by the correspondence function $t^{(m)} \colon \mathcal{V}^{(m)} \to \mathcal{X}$ that assigns the vertices of the $m$th 3D view to the corresponding vertices of the 3D shape.
\begin{figure}[t]
	\centering
	\begin{overpic}[width=\textwidth]{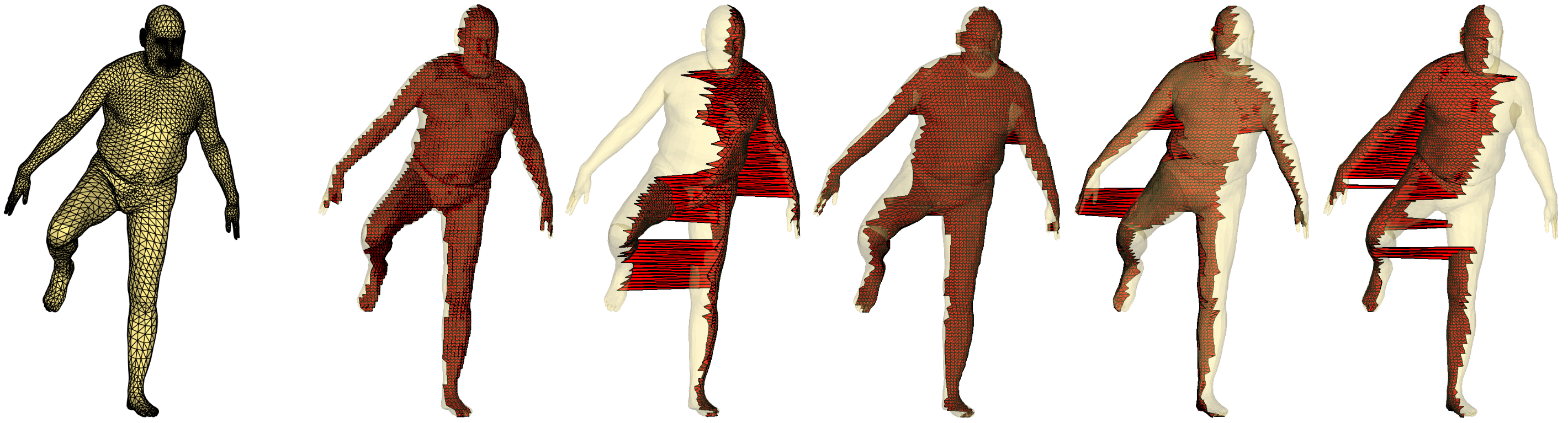} 
	    \put(18, -2){\rule{0.5pt}{35mm}}
    \end{overpic}
	\caption{Example of 3D augmented views. Left-hand side: a synthetic 3D shape from the FAUST dataset \cite{Bogo2014}. Right-hand side: examples of the 3D augmented views extracted by the shape decomposition module. 3D views have an uniform vertex density and capture the underlying geometry even at a lower resolution.}
	\label{fig:3dviews}
\end{figure}

\subsection{Feature extraction and classification}

The feature extraction and classification module processes the $M$ 3D views in parallel to learn features through a set of deep neural networks, namely ViewNets, with shared weights. 
Formally, each ViewNet is a non-linear parametric function $\boldsymbol{g}(\boldsymbol{v}_n) = \boldsymbol{\Phi}_{\boldsymbol{\Theta}_{\text{cla}}} \left( \boldsymbol{f}(\boldsymbol{v}_n) \right)$ that takes vertex-wise features $\boldsymbol{f}(\boldsymbol{v}_n)$ as input and produces the probability distribution $\boldsymbol{g}(\boldsymbol{v}_n) = (g_1(\boldsymbol{v}_n), \dots, g_L(\boldsymbol{v}_n))$ as output, where $L$ is the number of segmentation classes and $\boldsymbol{\Theta}_{\text{cla}} \subset \boldsymbol{\Theta}$ is the set of ViewNet learnable weights.
Let $\boldsymbol{g}_{\mathcal{V}^{(m)}} \in {[0, 1]}^{N^{(m)} \times L}$ be the matrix containing the pdfs of all vertices of $\mathcal{V}^{(m)}$.

A ViewNet module is defined as the composition of \emph{Intrinsic Convolutional} (IC), \emph{Fully Connected} (FC) and \emph{Softmax} layers. FC and Softmax are standard layers, whereas the IC layer replaces the convolutional layer used in traditional Euclidean CNNs to perform convolution operations on 3D views \cite{Monti2017}.
The convolution at $x \in \mathcal{X}$ using IC layers requires additional information, in the form of a local coordinate frame and a set of weighting functions that maps the signal of the local neighbourhood of $x$ to a fixed grid.

\subsection{Prediction aggregation}
Predictions inferred from each 3D view are re-projected and aggregated on the 3D shape $\mathcal{X}$ in order to transfer the segmentation result on the original input. We name this operation ProjNet.
ProjNet employs a pooling operation that takes the ViewNet predictions $\boldsymbol{g}_{\mathcal{V}^{(m)}}$ on $\mathcal{V}^{(m)}$ as input and the correspondence function $t^{(m)} \colon \mathcal{V}^{(m)} \to \mathcal{X}$ for any $m$, to produce a single confidence map $\boldsymbol{g}_{\mathcal{X}}$ defined on $\mathcal{X}$.
The pooling operation is defined as
\begin{equation*}
\boldsymbol{g}_{\mathcal{X}}(\boldsymbol{x}_n) = \frac{1}{\lvert \Omega(n) \rvert} \sum_{\tilde{m} \in \Omega(n)} \boldsymbol{g}_{\mathcal{V}^{(\tilde{m})}}(\boldsymbol{v}_{\tilde{n}}),
\end{equation*}
where $\Omega(n) = \{m\,:\,t^{(m)}(\tilde{n})=n\}$ is the set of 3D view indices relative to the vertex $\boldsymbol{x}_n \in \mathcal{X}$, and $\boldsymbol{g}_{\mathcal{V}^{(\tilde{m})}}(\boldsymbol{v}_{\tilde{n}})$ is the probability distribution over the segmentation classes associated to vertex $\boldsymbol{v}_{\tilde{n}}$ of the $\tilde{m}$th 3D view.

\subsection{Prediction refinement}

The output of ProjNet is a point-wise prediction, i.e. the label prediction of each vertex is estimated independently from its neighbors, thus leading to likely local label inconsistencies.
Moreover, some vertices of the input 3D shape may not have been projected on any of the 3D views, thus leading to vertices with undefined label predictions on $\mathcal{X}$.
Therefore, we impose local label consistency by using a surface-based Conditional Random Field (CRF) approach \cite{Kalogerakis2017,Zheng2015} that exploits contextual information to produce structured and dense predictions.

For each vertex $\boldsymbol{x}_n \in \boldsymbol{X}$, let $y_n \colon \boldsymbol{x}_n \to \mathcal{L}$ be a random variable that assigns a label $\ell \in \mathcal{L}$ to it, and let $\boldsymbol{y} = (y_1, \dots, y_N)$ be the set of the random variables associated to the $N$ vertices of $\boldsymbol{X}$.
The CRF energy associated to $\boldsymbol{y}$ is defined as:
\begin{equation}\label{eq:crf_energy}
E(\boldsymbol{y}) = \sum_{n=1}^N \psi_\textrm{unary}(y_n) + \sum_{n=1}^N \sum_{\tilde{n}=n+1}^N \psi_\textrm{pairwise}(y_n, y_{\tilde{n}}),
\end{equation}
where the unary term $\psi_{\textrm{unary}}(y_n)$ quantifies the assignment cost of $y_n$ to vertex $\boldsymbol{x}_n$ and the pairwise term $\psi_{\textrm{pairwise}}(y_n, y_{\tilde{n}})$ quantifies the joint assignment cost of $y_n, y_{\tilde{n}}$ to vertices $\boldsymbol{x}_n, \boldsymbol{x}_{\tilde{n}}$ \cite{Krahenbuhl2013}.
Because $\boldsymbol{g}_\mathcal{X}(\boldsymbol{x}_n)$ measures the cost of assigning the vertex $\boldsymbol{x}_n$ to $\mathcal{L}$, we define the unary term as $\psi_{\textrm{unary}}(y_n) = - \log \left( \boldsymbol{g}_\mathcal{X}(\boldsymbol{x}_n) \right)$.
The pairwise potential is instead defined as the weighted sum of three Gaussian kernels:
\begin{equation*}
    \begin{aligned}
    \psi_\textrm{pairwise}(y_n, y_{\tilde{n}}) &= \mu(y_n, y_{\tilde{n}}) \bigg( w_{\textrm{near}}\; k_{\textrm{near}}(y_n, y_{\tilde{n}}) - w_{\textrm{far}}\; k_{\textrm{far}}(y_n, y_{\tilde{n}})\\ &+  w_{\textrm{feat}}\; k_{\textrm{feat}}(y_n, y_{\tilde{n}}) \bigg),
    \end{aligned}
\end{equation*}
where 
\begin{align*}
    &k_{\textrm{near}}(y_n, y_{\tilde{n}}) = \exp \left( -\frac{ d_{\mathcal{X}}(\boldsymbol{x}_n, \boldsymbol{x}_{\tilde{n}}) }{\sigma_{\textrm{near}}} \right), & \nonumber\\
    &k_{\textrm{far}}(y_n, y_{\tilde{n}}) = 1_{\mathcal{X}} - \exp \left( -\frac{ d_{\mathcal{X}}(\boldsymbol{x}_n, \boldsymbol{x}_{\tilde{n}}) }{ \sigma_{\textrm{far}} } \right), & \nonumber\\
    &k_{\textrm{feat}}(y_n, y_{\tilde{n}}) = \exp \left( -\frac{\lVert \boldsymbol{f}(\boldsymbol{x}_n) - \boldsymbol{f}(\boldsymbol{x}_{\tilde{n}}) \rVert_2}{\sigma_{\textrm{feat}}} \right), &
\end{align*}
$d_{\mathcal{X}}(x, \tilde{x})$ is the geodesic distance between the vertices $x, \tilde{x} \in \mathcal{X}$, $1_{\mathcal{X}}$ is the identity function on $\mathcal{X}$, and $\mu(y_n, y_{\tilde{n}})$ is a label compatibility term.

\begin{figure}[t]
	\centering
	\begin{overpic}[width=\textwidth]{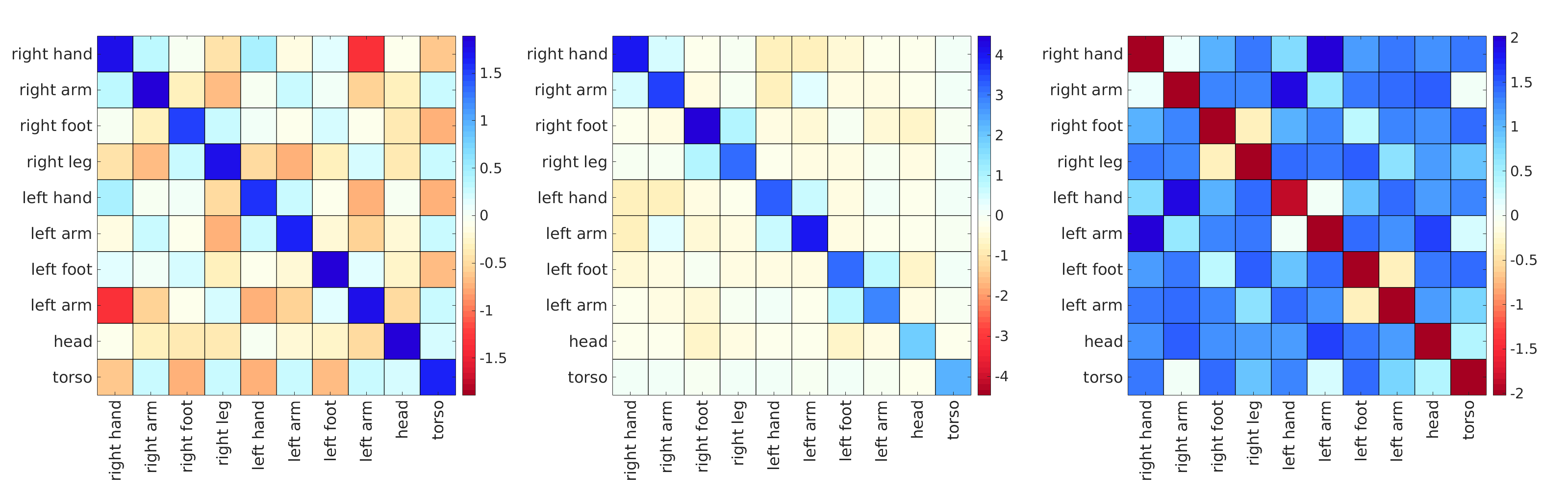} 
	    \put(14.5, 31.5){\footnotesize $w_{\textrm{near}}$}
	    \put(48.5, 31.5){\footnotesize $w_{\textrm{far}}$}
	    \put(83.5, 31.5){\footnotesize $\mu$}
    \end{overpic}
    \vspace{-.7cm}
	\caption{Example of Conditional Random Field (CRF) learned weights ($w_{\textrm{near}}, w_{\textrm{far}}, \mu$) in the case of human 3D shapes.}
	\label{fig:crf_kernels}
\end{figure}

Similarly to \cite{Kalogerakis2017,Zheng2015},  $k_{\textrm{near}}$ favors local spatial consistency, while $k_{\textrm{feat}}$ promotes the assignment of similar labels to vertices with similar properties.
The third kernel $k_{\textrm{far}}$ is novel and is introduced to disambiguate symmetries. Because symmetric parts are likely to be located far from each other (e.g.~arms and legs in a human shape) we designed $k_{\textrm{far}}$ to avoid distant points to have similar labels.
The set of CRF learnable parameters is defined as $\boldsymbol{\Theta}_{\text{CRF}} = \{ \mu, w_{\textrm{near}}, w_{\textrm{far}}, w_{\textrm{feat}}\}$, $\boldsymbol{\Theta}_{\text{CRF}} \subset \boldsymbol{\Theta}$.
Fig.~\ref{fig:crf_kernels} shows how CRF learns the relationships among segments through an example of learned parameters (i.e.~$w_{\textrm{near}}$, $w_{\textrm{far}}$ and $\mu$) on human 3D shapes. In $w_{\textrm{near}}$ we can observe that the head weights suggest that there is a strong relationship between head and torso rather than between head and right foot/right arm. Similarly, the torso weights suggest that there is a strong relationship between torso and arms/legs rather than between torso and feet/hands.

The most probable pdf configuration of $\boldsymbol{y}$ for $\mathcal{X}$ is obtained by minimizing the energy $E(\boldsymbol{y})$ defined in Eq.~\ref{eq:crf_energy}. The exact inference of the CRF distribution is intractable, thus we use a mean-field approximation \cite{Kalogerakis2017,Krahenbuhl2013}. The iterative algorithm for approximate mean-field inference can be implemented as a Recurrent Neural Network (RNN) by rephrasing each step of the algorithm as a CNN layer \cite{Zheng2015}.

\section{Results}

\subsection{Experimental setup}

We evaluate our 3D segmentation approach through two different experiments.
Firstly, we use data from the publicly available Princeton Shape Benchmark (PSB) dataset \cite{Shilane2004} that contains synthetic shapes of several objects and animals; in particular, the rigid shapes of the Airplane class, and the non-rigid shapes of the Ant, Four Leg and Teddy classes. The segmentation labels of each object are defined as in \cite{Shilane2004}.
Secondly, we use data of non-rigid human shapes; in particular, (i) synthetic people with different poses (FAUST dataset \cite{Bogo2014}), (ii) real people acquired with depth sensors (SCAPE dataset \cite{Anguelov2005}) and with structured light 3D body scanners (SHREC14 dataset \cite{Pickup2014}), and (iii) real people that we acquired with a smartphone and reconstructed using the photogrammetry pipeline COLMAP \cite{Schonberger2016}.
We manually labelled the ground truth for FAUST and SCAPE datasets and used their training data to learn the neural network model for the human shapes.
We have used this model to test our approach on all the other human shapes of FAUST, SCAPE, SHREC14 and COLMAP datasets. The segmentation labels for the non-rigid human shapes are: $\mathcal{L} = \{$head, torso, right arm, right hand, right leg, right foot, left arm, left hand, left leg, left foot$\}$.

\subsection{Training}
Given a labelled training set, where each vertex $\boldsymbol{x}_n \in \boldsymbol{X}$ is associated to a ground-truth label $h(\boldsymbol{x}_n)$, the optimal parameters are obtained by minimizing the \emph{categorical cross-entropy} loss,
\begin{equation*}
c(\delta_{h(\boldsymbol{x}_n)}, \boldsymbol{\Gamma}_{\boldsymbol{\Theta}}(x)) = -\sum_{n=1}^N \delta_{h(\boldsymbol{x}_n)} \log( \boldsymbol{\Gamma}_{\boldsymbol{\Theta}}(\boldsymbol{x}_n) ),
\end{equation*}
where $\delta_{h(\boldsymbol{x}_n)}$ is the Kronecker delta defined for the ground-truth label $h(\boldsymbol{x}_n)$.

Our approach is trained end-to-end and from scratch. We use $M=10$ 3D views (Sec.~2.2, Fig.~\ref{fig:3dviews}) taken from equi-spaced viewpoints around the shape. For training we use the Adam optimizer \cite{Kingma2015} with a learning rate of $0.001$. The CRF weights are initialized with identity matrices, i.e.~each segment class is only in relationship with itself.

\subsection{Evaluation}
\begin{figure}[htb]
	\centering
	\begin{overpic}[width=\textwidth]{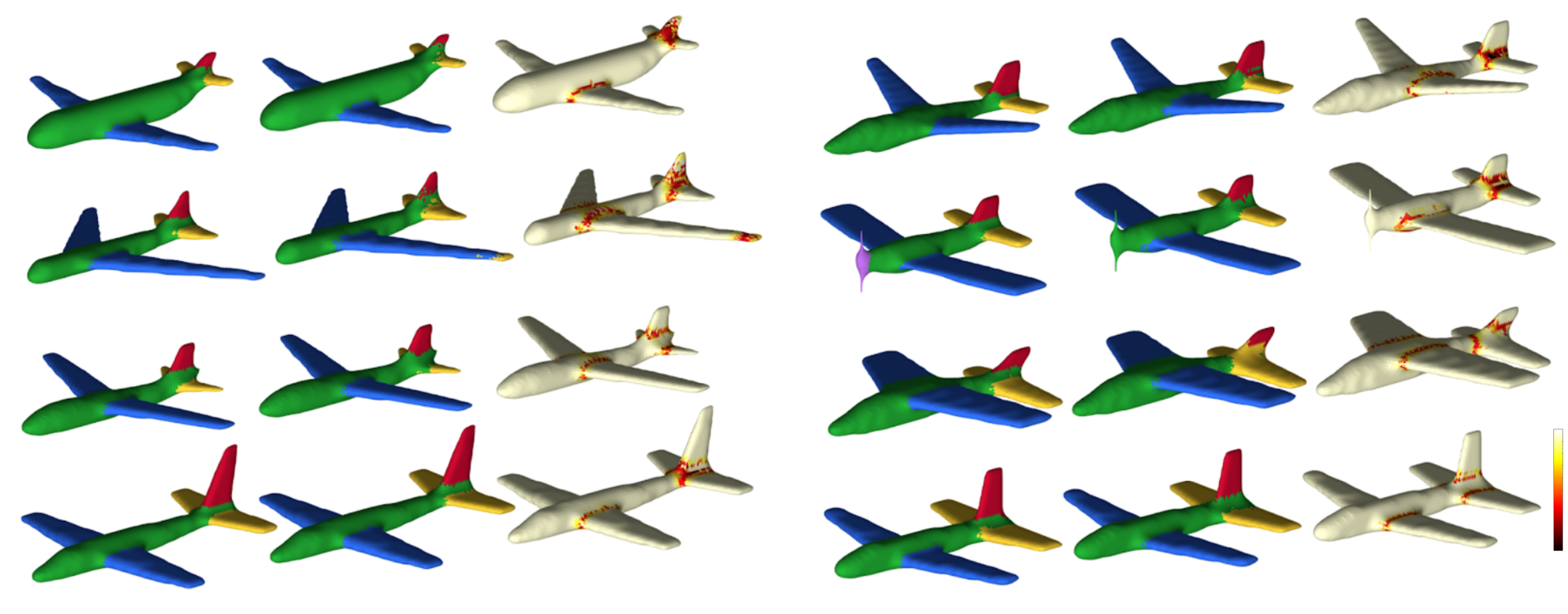}
        \put(100, 3.5){\footnotesize{$0$}}
        \put(100, 10.5){\footnotesize{$1$}}
    \end{overpic}
    \vspace{-.7cm}
	\caption{Semantic segmentation results of our approach on PSB Airplane test shapes. Segmentation color key: green = body, blue = wings, purple = engine, yellow = stabilizer, and red = rudder. Each segmentation result (center) is accompanied by its ground-truth (on its left) and a confidence map (on its right) showing the uncertainty (entropy) of the network prediction over the 3D shape. The darker the color the higher the uncertainty.}
	\label{fig:results_PSB}
\end{figure}

\noindent\textbf{PSB dataset:} Table \ref{tab:PSB_mAcc} shows the quantitative results of our approach on a subset of PSB's 3D shapes. We compare the accuracy of our approach with ShapeBoost \cite{Kalogerakis2010}, Guo et al.~\cite{Guo2015} and ShapePFCN \cite{Kalogerakis2017}. 
The first two approaches use classifiers that are learned from hand-crafted features, whereas the latter is an end-to-end deep learning approach similar to ours (i.e.~features are also learned).
We can observe that the accuracy of our approach is similar to that of state-of-the-art methods. However, compared to ShapePFCN \cite{Kalogerakis2017} that is based on the VGG16 architecture \cite{Simonyan2015}, which uses $134$M parameters, our neural network uses $14$K parameters, i.e.~$1\%$ of ShapePFCN's parameters \cite{Kalogerakis2017}.
Fig.~\ref{fig:results_PSB} shows examples of segmentation results that are obtained on the Airplane category. The uncertainty map next to each segmentation result showed that the highest level of uncertainty is located where different segments intersect. Qualitatively, the results are very accurate and show only minor errors on the rudder region.
\vspace{-3mm}
\begin{table*}[htb]
\caption{Segmentation mean accuracy (the higher the better \cite{Kalogerakis2017}) on the Princeton Shape Benchmark dataset \cite{Shilane2004}.}
\label{tab:PSB_mAcc}
\centering
\tabcolsep 4pt
\begin{tabular}{l|cc|cc}
    \multirow{2}{*}{Category} & %
    \multicolumn{2}{c|}{Hand-crafted features} & %
    \multicolumn{2}{c}{End-to-end} \\
    & 
    ShapeBoost \cite{Kalogerakis2010} & %
    Guo et al. \cite{Guo2015} & %
    ShapePFCN \cite{Kalogerakis2017} & %
    Ours\\ 
    \hline
    Airplane & 96.1 & 91.6 & 93.0 & 94.1\\
    Ant      & 98.7 & 97.6 & 98.6 & 94.8\\
    Four Leg & 83.3 & 82.4 & 85.0 & 94.5\\
    Teddy    & 98.7 & 97.3 & 97.7 & 92.8\\
    \hline
\end{tabular}
\end{table*}

\noindent\textbf{Non-rigid human shapes:} Fig.~\ref{fig:results_FAUST} to \ref{fig:results_REPLICATE} show examples of segmentation results that are obtained on the non-rigid human shapes. Beside each segmented shape we can observe their associated entropy map. The smaller the entropy the higher the uncertainty. As expected, the largest level of uncertainty is located at the joints between two segments, that is where transition is not well defined.
Because we have annotations for FAUST and SCAPE, we quantified the accuracy \cite{Kalogerakis2017} and Intersection over Union (IoU) \cite{Qi2017} of the segmentation results. In FAUST we achieved an accuracy of 93.8\% and IoU of 88.5\% while in SCAPE we achieved an accuracy of 72.1\% and IoU of 58.7\%. This accuracy and IoU differences are due to the unbalanced number of training samples of the two datasets. FAUST annotations are much more numerous than those of SCAPE. A few of the poses of FAUST's training shapes are also present in the test set. This does not occur in the case of SCAPE, where poses are only present once. Fig.~\ref{fig:results_SCAPE} shows examples of the segmentation errors occurred in SCAPE test, e.g.~on the right-hand block we can see that the legs of the shape in the middle have been segmented with inverted labels. 
\begin{figure}[htb]
	\centering
	\begin{overpic}[width=\textwidth]{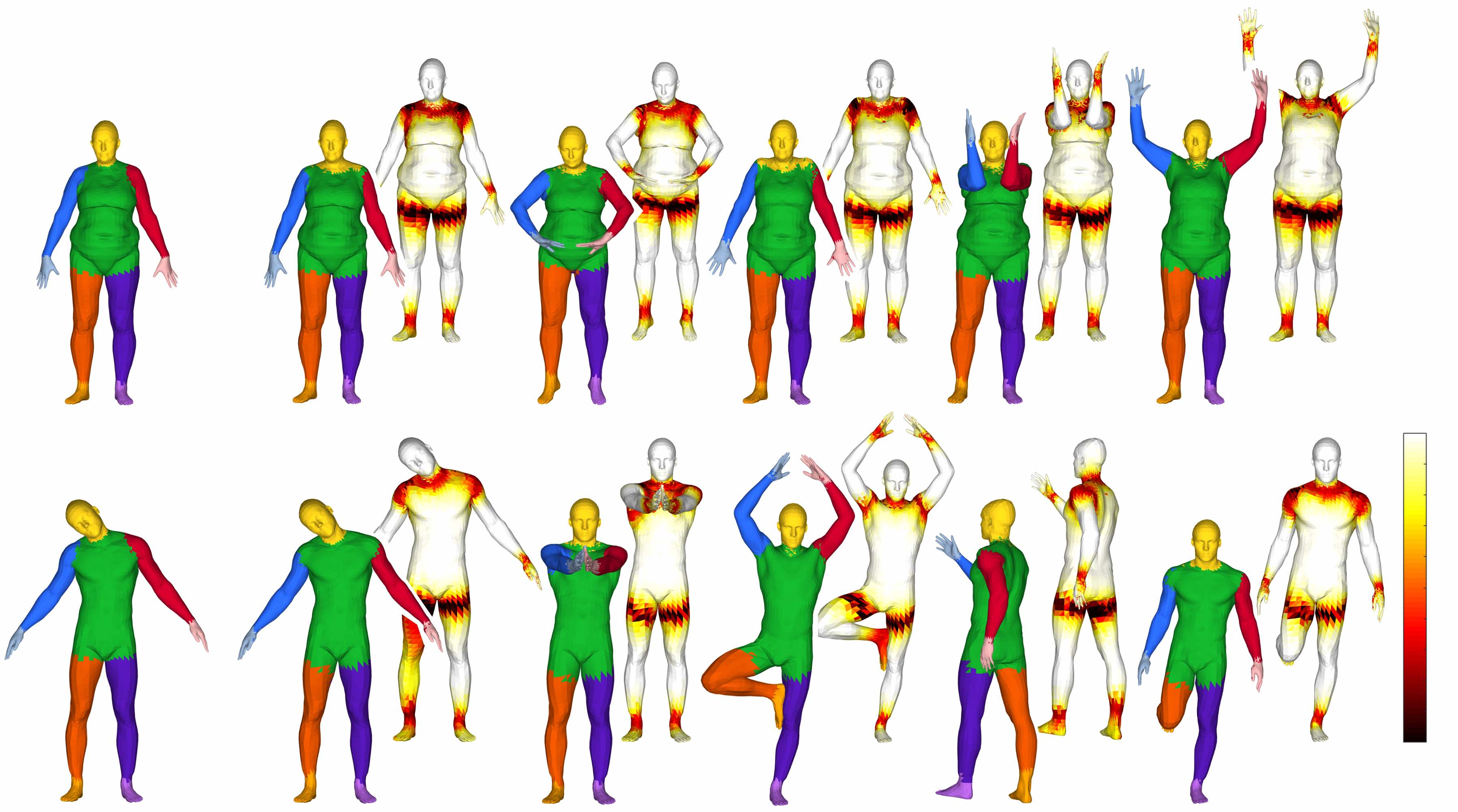}
        \put(15.5, -3){\rule{0.5pt}{71mm}}
        \put(5, -2.5){\footnotesize{GT}}
        \put(20, -2.5){\footnotesize{preds}}
        \put(27, 2){\footnotesize{confs}}
        \put(98.25, 4){\footnotesize{$0$}}
        \put(98.25, 25){\footnotesize{$1$}}
    \end{overpic}
	\caption{Semantic segmentation results of our approach on a subset of FAUST's test shapes. Segmentation color key: colour code: yellow = head, green = torso, blue = right arm, light blue = right hand, orange = right leg, yellow = right foot, red = left arm, light red = left hand, purple = left leg, light purple = left foot. Each segmentation result (left) is accompanied by a confidence map (right) showing the uncertainty (entropy) of the network prediction over the 3D shape. The darker the color the higher the uncertainty.}
	\label{fig:results_FAUST}
\end{figure}
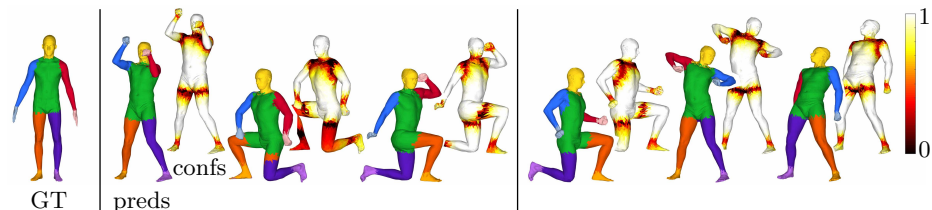
\begin{figure}[htb]
	\centering
	\begin{overpic}[width=\textwidth]{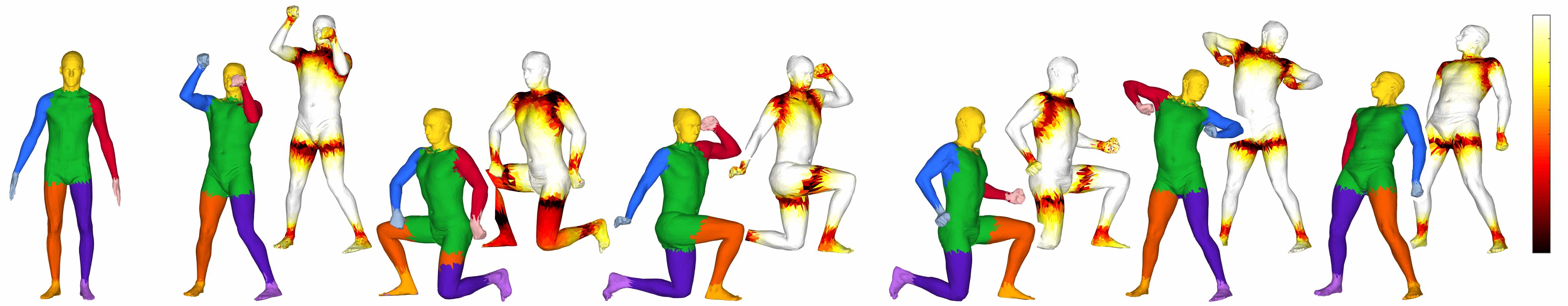} 
        \put(10, -3){\rule{0.5pt}{27mm}}
        \put(55.5, -3){\rule{0.5pt}{27mm}}
        \put(2.5, -2.5){\footnotesize{GT}}
        \put(11.5, -2.5){\footnotesize{preds}}
        \put(18, 1){\footnotesize{confs}}
        \put(99.25, 3){\footnotesize{$0$}}
        \put(99.25, 17.5){\footnotesize{$1$}}
    \end{overpic}
	\caption{Semantic segmentation results of our approach on a subset of SCAPE's test shapes. Segmentation color key is the same as that in Fig.~\ref{fig:results_FAUST}. 
	}
	\label{fig:results_SCAPE}
\end{figure}

Results in Fig.~\ref{fig:results_SHREC14} and \ref{fig:results_REPLICATE} show that the method can generalize also to 3D shapes that have not been used for training. Interestingly, our approach can effectively generalize the mesh representation through the 3D augmented views and produce a reliable segmentation in the case of COLMAP's shapes. Note that the mesh topology of COLMAP's shapes is different from those used in training. This is because the meshing operation based on Poisson reconstruction of COLMAP produces highly irregular polygons \cite{Kazhdan2006}. However, it is also clear that COLMAP's shapes are more challenging than SHREC14's ones by looking at the respective confidence maps. Overall, results show that our approach can effectively segment 3D shapes of different subjects, despite their different pose.
\begin{figure}[t]
	\centering
	\begin{overpic}[width=\textwidth]{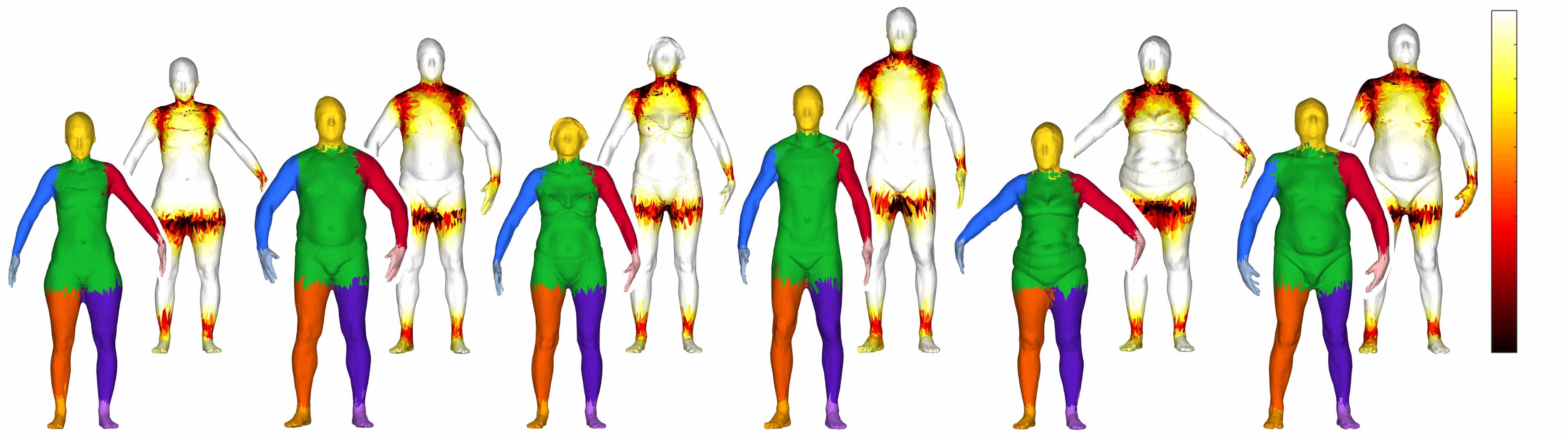} 
        \put(2, -2){\footnotesize{preds}}
        \put(9, 2.5){\footnotesize{confs}}
        \put(97.5, 5){\footnotesize{$0$}}
        \put(97.5, 26){\footnotesize{$1$}}
    \end{overpic}
	\caption{Semantic segmentation results of our approach on a subset of SHREC14's shapes. Trained on FAUST and SCAPE training sets. Segmentation color key is the same as that in Fig.~\ref{fig:results_FAUST}. 
	}
	\label{fig:results_SHREC14}
\end{figure}
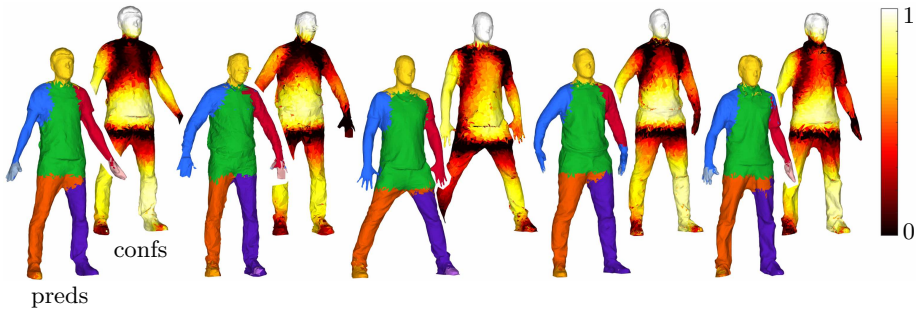
\begin{figure}[t]
	\centering
	\begin{overpic}[width=\textwidth]{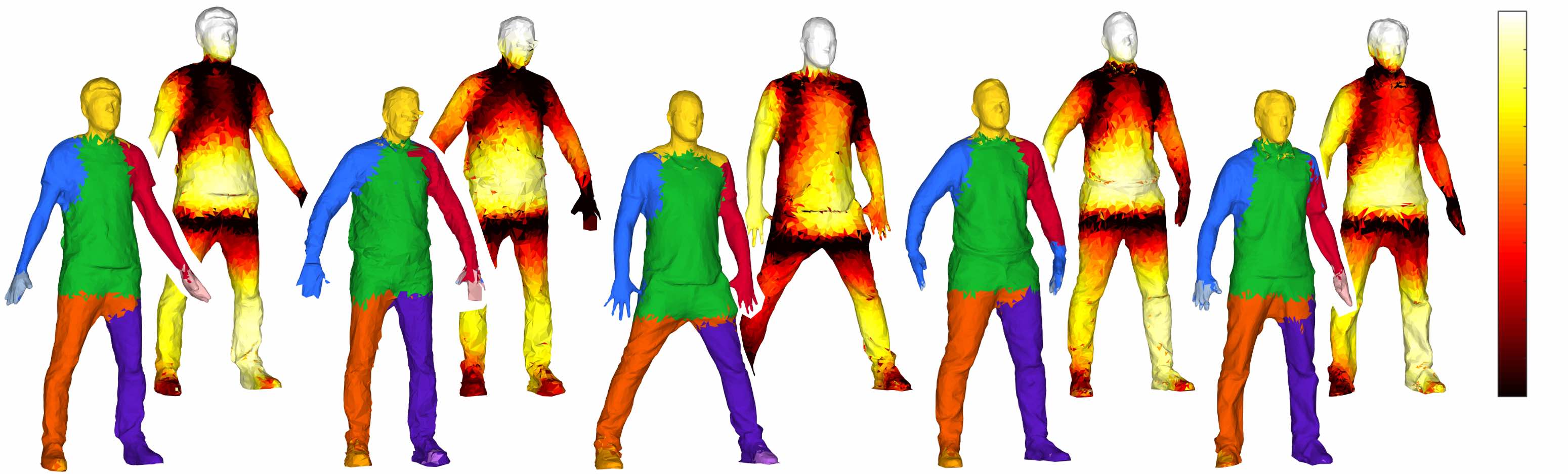}
        \put(3, -2.5){\footnotesize{preds}}
        \put(12, 2.5){\footnotesize{confs}}
        \put(98, 4.5){\footnotesize{$0$}}
        \put(98, 28){\footnotesize{$1$}}
    \end{overpic}
	\caption{Semantic segmentation results of our approach on a subset of COLMAP's shapes. Trained on FAUST and SCAPE training sets. Segmentation color key is the same as that in Fig.~\ref{fig:results_FAUST}. 
	}
	\label{fig:results_REPLICATE}
\end{figure}

\section{Conclusions}

We presented an approach to segment 3D shapes efficiently regardless their mesh topology. To achieve this we decomposed the segmentation problem into sub-segmentation problems by using 3D augmented views generated from the underlying 3D shape. This enabled us to train a neural network with 1\% of the parameters used by alternative state-of-the-art solutions, while maintaining similar accuracy performance. We showed that our approach is generic and can be used to segment 3D shapes with arbitrary mesh topologies, like those computed with photogrammetry reconstruction techniques (e.g.~Poisson reconstruction \cite{Kazhdan2006}) that have a high density of polygons and that are distributed irregularly. Moreover, our approach also showed evidence of being flexible to segment other categories of 3D shapes (e.g.~airplanes) other than human ones.

Future research directions include an extensive analysis of the results, evaluating the impact of a multi-scale approach applied on the 3D augmented views and exploring next-best-view approaches \cite{Potthast2014} to select the most suitable 3D views of the object of interest. We will also exploit the structured output of the CRF to build models for surface matching between 3D shapes \cite{Boscaini2016} and explore attention mechanisms to make the prediction of our approach robust to the clutter present on the 3D shape (i.e.~untrained segmentation classes).

\bibliographystyle{splncs04}
\bibliography{refs}

\end{document}